\pdfoutput=1

\documentclass[11pt]{article}
\usepackage{authblk}
\usepackage{enumitem}
\usepackage{EMNLP2023}
\usepackage{comment}
\usepackage{times}
\usepackage{booktabs}
\usepackage{multirow}
\usepackage{latexsym}
\usepackage{mathtools}
\usepackage{tikz}
\usepackage{graphicx}
\usepackage{multicol}
\usetikzlibrary{positioning,calc,arrows.meta}

\usepackage[T1]{fontenc}
\usepackage[utf8]{inputenc}

\newcommand{\Xianzhi}[1]{\textcolor{red}{#1}}

\title{Are \texttt{ChatGPT} and \texttt{GPT-4} General-Purpose Solvers for \\  Financial Text Analytics? A Study on Several Typical Tasks}

\author{Xianzhi Li\textsuperscript{1}, Samuel Chan\textsuperscript{1}, Xiaodan Zhu\textsuperscript{1}, Yulong Pei\textsuperscript{2}, Zhiqiang Ma\textsuperscript{2}, Xiaomo Liu\textsuperscript{2} and Sameena Shah\textsuperscript{2}\\
\textsuperscript{1}Department of Electrical and Computer Engineering \& Ingenuity Labs Research Institute \\ Queen's University\\
\textsuperscript{2}J.P. Morgan AI Research\\
\small{\{\texttt{li.xianzhi, 19syc2, xiaodan.zhu}}\}\text{\texttt{@queensu.ca}} \\\small{\{\texttt{yulong.pei,zhiqiang.ma,xiaomo.liu,sameena.shah}}\}\text{\texttt{@jpmchase.com}}
}

\begin{document}

\maketitle

\begin{abstract}
The most recent large language models (LLMs) such as \texttt{ChatGPT} and \texttt{GPT-4} have shown  exceptional capabilities of generalist models, achieving state-of-the-art performance on a wide range of NLP tasks with little or no adaptation. How effective are such models in the financial domain? Understanding this basic question would have a significant impact on many downstream financial analytical tasks. In this paper, we conduct an empirical study and provide experimental evidences of their performance on a wide variety of financial text analytical problems, using eight benchmark datasets from five categories of tasks. We report both the strengths and limitations of the current models by comparing them to the state-of-the-art fine-tuned approaches and the recently released domain-specific pretrained models. We hope our study can help understand the capability of the existing models in the financial domain and facilitate further improvements.

\end{abstract}

\section{Introduction} 
The advancement of LLMs is bringing profound impacts on the financial industry. Through training with reinforcement learning from human feedback (RLHF) \citep{christiano:23} and masked language model objectives, the most recent models such as \texttt{ChatGPT}\footnote{https://platform.openai.com/docs/models/gpt-3-5} and \texttt{GPT-4}\footnote{https://platform.openai.com/docs/models/gpt-4} have demonstrated exceptional capabilities in a wide range of natural language processing (NLP) tasks~\cite{bang2023multitask,liu2023evaluating,omar2023chatgpt,khoury2023secure}. 

These LLMs are trained on datasets that encompass a broad range of genres and topics. While their performance in generic NLP tasks is impressive, their applicability and effectiveness in specific domains like finance yet need a better understanding and can influence a wide range of applications. In general, in the financial domain, LLMs is playing an increasingly crucial role in tasks such as investment sentiment analysis, financial named entity recognition, and question-answering systems for assisting financial analysts.

In this paper, we perform an empirical study and provide experimental evidence for the effectiveness of the most recent LLMs on a variety of financial text analytical problems, involving eight benchmark datasets from five typical tasks. These datasets are from a range of financial topics and sub-domains such as stock market analysis, financial news, and investment strategies. We report both the strengths and limitations of \texttt{ChatGPT} and \texttt{GPT-4} by comparing them with the state-of-the-art domain-specific fine-tuned models in finance, e.g., \texttt{FinBert} \citep{araci2019finbert} and \texttt{FinQANet} \citep{chen2022finqa}, as well as the recently pretrained model such as \texttt{BloombergGPT} \citep{wu2023bloomberggpt}.
Our main contributions are summarized as follows:
\vspace{1mm}
\begin{itemize}[leftmargin=5mm,itemsep=1mm,parsep=1mm]
\item This study is among the first to explore the most recent advancement of \textit{generically} trained large language models on financial text analytic tasks and it provides a comprehensive comparison.
\item We demonstrate that \texttt{ChatGPT} and \texttt{GPT-4} can outperform the most recently released domain-specifically pretrained model as well as 
 fine-tuned models on many tasks. We provide detailed analysis and recommendations.
\item We observe that the advancement made in generalist models continues to carry over to the financial domain; e.g., \texttt{GPT-4} is significantly better than \texttt{ChatGPT} on nearly all the financial benchmarks used. 
\item Limitations of the existing LLMs are analyzed and discussed with these benchmark datasets.
\begin{table*}[htbp]
\centering
\resizebox{\textwidth}{1.5cm}{%
\begin{tabular}{c||c|c|c|c|c}
\hline
\textbf{Category} & \textbf{Sentiment Analysis} & \textbf{Classification} & \textbf{NER}  & \textbf{RE}  &\textbf{QA} \\ \hline
\textbf{Complexity} &Easy &Easy &Hard &Hard &Hard \\ \hline
\textbf{Knowledge}  &Low  &Low  &High &High &High \\ \hline
\textbf{Dataset} & FPB/FiQA/TweetFinSent  & Headline & NER &  REFinD & FinQA/ConvFinQA\\ \hline
\textbf{Eval. Metrics} & Weighted F1 & Weighted F1 & Macro F1 & Macro F1 & Accuracy \\ \hline
\textbf{\#Test samples} & 970/223/996 & 2,114 & 98 & 4300 & 1,147/421\\ \hline
\end{tabular}%
}
\caption{Statistics of the five tasks and eight datasets used in this study.}
\label{tab:data}
\end{table*}

\end{itemize}

\newpage
\section{Related Works} 
\label{sec:rw}
\paragraph{\texttt{ChatGPT} and Related Models.}
\texttt{ChatGPT}, \texttt{GPT-3.5} (\texttt{text-davinci-003}), and \texttt{GPT-4} are generically trained LLMs and have shown impressive performance on a wide range of tasks. Recent studies have shown that they outperform fine-tuned models on some tasks. But, they still fail in some other cases. \citet{bang:23} evaluated \texttt{ChatGPT} on multitasking, multilingual and multimodal tasks, highlighting addressing the failures  to improve the overall performance. \citet{qin:23} studied \texttt{ChatGPT}'s zero-shot capabilities on a diverse range of NLP tasks. While these models present unprecedented quality and retain accumulated knowledge with excellent generalization ability, by respecting the objective of being general problem solvers, how effective they are for financial text analytical tasks is an intriguing open question that needs a better understanding. 

\paragraph{Domain-specific Models}
Currently, there have been only a handful of LLMs specifically trained within the finance domain. \texttt{BloombergGPT} \citep{wu2023bloomberggpt}, a language model with 50 billion parameters, is trained using a mixed approach to cater to the financial industry's diverse tasks. The model is evaluated on standard LLM benchmarks, open financial benchmarks, and Bloomberg-internal benchmarks. The mixed training approach results in a model that significantly outperforms existing models in financial tasks and performs on par or even better in some general NLP benchmarks.
Other researchers also attempted to adapt existing language models to tackle domain-specific tasks. For example, 
~\citet{lewkowycz2022solving} adapted \texttt{T5} to the financial domain. Note that in addition to fine-tuning, A study has also been conducted to use parameter-efficient tuning for financial tasks such as intent detection~\cite{li2022learning}. The details of the related work can be found in Appendix~\ref{app:related_work}. 

\section{Experiment Setup} 
\label{sec:exp}
\paragraph{Tasks and Datasets.}
Our research utilizes a wide range of financial NLP tasks and challenges \cite{pei-etal-2022-tweetfinsent, kaur2023refind,shah2022flue}, enabling us to establish a testbed with different types of NLP problems ranging from basic sentiment analysis and text classification to information extraction and  question answering (see Table \ref{tab:data} and more details in Appendix \ref{app:dataset_details}).  

The span of the tasks enables us to make observations along modeling complexity and different levels of financial knowledge required to perform the tasks. Regarding the modeling complexity of tasks, sentiment analysis and text classification are often regarded to be more straightforward, compared to information extraction (IE) tasks such as named entity recognition (NER) and relation extraction (RE). The latter often requires more understanding of syntax and semantics in the input contexts as well as the interactions of labels in the output space as the \textit{structured prediction} problems. Compared to sentiment analysis and text classification, question answering (QA) is often thought of as being harder as it often requires a model to understand the embedded internal logic and numerical operation/reasoning. Regarding financial knowledge, the existing classification and sentiment analysis datasets are sourced from daily news and social media. On the other hand, IE and QA data are often from professional documents like financial filings and reports, which usually require more domain knowledge to comprehend.


\paragraph{Models.}
We test the representative state-of-the-art LLMs, \texttt{ChatGPT} and \texttt{GPT-4} models. Specifically, we use \texttt{gpt-3.5-turbo} and \texttt{GPT-4 (8k)} for most of the experiments, except FinQA few-shot experiments, where the input tokens are extra long so we adopt \texttt{gpt-3.5-turbo-16k}.\footnote{All the models are current versions as of July 7th, 2023. } Both these LLMs are evaluated using zero-shot and few-shot learning as well as CoT learning for QA reasoning tasks. Furthermore, we compare them with previous LLMs and the domain specific \texttt{BloombergGPT} \cite{wu2023bloomberggpt}. The state-of-the-art fine-tuned models on each dataset are employed to test the idea of training smaller models on individual tasks in comparison with prompting LLMs on all tasks without additional fine-tuning.

\paragraph{Evaluation Metrics.}
We use \textit{accuracy}, \textit{macro-F1 score}, and \textit{weighted F1 score} \citep{wu2023bloomberggpt} as the evaluation metrics. For the NER task, we calculate the \textit{entity-level F1 score}.
Table~\ref{tab:data} shows the details of the experiment setup.

\section{Results and Analysis}

\subsection{Sentiment Analysis}
Sentiment analysis is one of the most commonly used NLP techniques in the financial sector and can be used to predict investment behaviors and trends in equity markets from news and social media data \cite{mishev2020evaluation}. We use three financial sentiment datasets with different focuses. 

\paragraph{Financial PhraseBank.} PhraseBank is a typical three scale (positive, negative and neutral) sentiment classification task curated from financial news by 5-8 annotators \citep{malo2013good}. We use both the \textit{50\% annotation agreement} and the \textit{100\% agreement} datasets. Same as in~\citep{wu2023bloomberggpt}, 20\% sentences are used for testing. In Table~\ref{table1}, the first group  of models (4 models) are OpenAI LLMs, followed by \texttt{BloombergGPT}, three previous LLMs (referred to as \texttt{Prior LLMs}), and the state-of-the-art fine-tuned models on this dataset (\texttt{FinBert}). Due to the space limit of Table~\ref{table1}, we put the name of these four groups in the next table (Table ~\ref{tab:FiQA}) for clarity. In Table~\ref{table1}, we can see that the performance of \texttt{Prior LLMs} greatly falls behind \texttt{ChatGPT} and \texttt{GPT-4}. With the enhancement of few-shot learning, \texttt{GPT-4} is comparable to fine-tuned \texttt{FinBert} \citep{araci2019finbert}.  

\begin{table}[htbp]
\centering
\resizebox{\columnwidth}{!}{%
\begin{tabular}{l|c|c|c|c}
\hline
Data & \multicolumn{2}{|c|}{50\% Agreement} & \multicolumn{2}{|c}{100\% Agreement} \\ \hline
Model           & Accuracy  & F1 score  & Accuracy & F1 score  \\ \hline
\texttt{ChatGPT{\scriptsize (0)}}         & 0.78      & 0.78      & 0.90    & 0.90      \\
\texttt{ChatGPT{\scriptsize (5)}}         & 0.79      & 0.79      & 0.90    & 0.90      \\
\texttt{GPT-4{\scriptsize (0)}}           & \underline{0.83}      & \underline{0.83}      & \underline{0.96}     & \underline{0.96}      \\ 
\texttt{GPT-4{\scriptsize (5)}}           & \textbf{0.86}      & \textbf{0.86}      & \textbf{0.97}     & \textbf{0.97}      \\ \hline

\texttt{BloombergGPT{\scriptsize (5)}}   & /         & 0.51      & /        & /         \\ \hline
\texttt{GPT-NeoX{\scriptsize (5)}}        & /         & 0.45      & /        & /         \\
\texttt{OPT66B{\scriptsize (5)}}          & /         & 0.49      & /        & /         \\
\texttt{BLOOM176B{\scriptsize (5)}}       & /         & 0.50      & /        & /         \\ \hline
\texttt{FinBert}         & \textbf{0.86}      & 0.84      & \textbf{0.97}     & 0.95      \\ \hline
\end{tabular}%
}
\caption{Results on the Phrasebank dataset. The subscript $(n)$ after an LLM name represents the number of shots. The best results are marked in bold and the second-best with underscored. The results of other LLMs like \texttt{BloombergGPT} are from the corresponding papers. `/' indicates the metrics were not included in the original study. The notation convention used here applies to all the following experiments. Different few-shot settings are tested and discussed in Appendix \ref{app:fewshot}.}
\label{table1}
\end{table}

\paragraph{FiQA Sentiment Analysis.} This dataset extends the task complexity to detect aspect-based sentiments from news and microblog in the financial domain \cite{10.1145/3184558.3192301}. We follow \texttt{BloombergGPT}'s setting \citep{wu2023bloomberggpt}, where we cast this regression task into a classification task. 20\% of labeled training data are held as test cases. 
The results in Table~\ref{tab:FiQA} present similar performance trends as in the previous dataset: \texttt{ChatGPT} and \texttt{GPT-4} outperform \texttt{Prior LLMs}. With a few-shot examples \texttt{GPT-4} is better than all other models here. \texttt{BloombergGPT} has relatively close performance to zero-shot \texttt{ChatGPT} and is inferior to \texttt{GPT-4}. The fine-tuned \texttt{RoBERTa-large} model on this dataset is better than \texttt{ChatGPT}, but is slightly less effective than \texttt{GPT-4}. The latter achieves 88\% on F1, which is less than that in Financial PhraseBank. We due this to the fact that FiQA requires modeling more details and needs more domain knowledge to understand the sentiment with the aspect finance tree in the data.


\begin{table}[htbp]
\fontsize{10pt}{10pt}\selectfont
\centering
\begin{tabular}{l|c|c}
\hline
\textbf{Model}   & \textbf{Category}      & \textbf{Weighted F1} \\
\hline
\texttt{ChatGPT{\scriptsize (0)}}    & \multirow{4}{*}{\shortstack{OpenAI\\LLMs}} &   75.90     \\
\texttt{ChatGPT{\scriptsize (5)}}   &     & 78.33      \\
\texttt{GPT-4{\scriptsize (0)}}      &    & \underline{87.15}       \\
\texttt{GPT-4{\scriptsize (5)}}      &    & \textbf{88.11}       \\ \hline
\multirow{2}{*}{\texttt{BloombergGPT{\scriptsize (5)}}} & Domain & \multirow{2}{*}{75.07} \\
&LLM & \\ \hline
\texttt{GPT-NeoX{\scriptsize (5)}}  & \multirow{3}{*}{\shortstack{Prior\\LLMs}}     & 50.59       \\
\texttt{OPT66B{\scriptsize (5)}}   &      & 51.60       \\
\texttt{BLOOM176B{\scriptsize (5)}}  &    & 53.12       \\ \hline
\texttt{RoBERTa-large}      & Fine-tune               & 87.09 \\ 
\hline
\end{tabular}
\caption{Results on the  FiQA dataset.}
\label{tab:FiQA}
\end{table}

\paragraph{TweetFinSent.} \citet{pei-etal-2022-tweetfinsent} created this dataset based on Twitter to capture retail investors' mood to a specific stock ticker. Since tweets are informal texts which typically are not used to train LLMs, this could be a challenging task for LLMs to perform well. Furthermore, a tweet can sometimes contain several tickers (>5 is not unusual). The aspect modeling on this data is more complex. The evaluation results on 996 test instances are shown in Table~\ref{tab:TweetFinSent}. \texttt{GPT-4} with a few-shot examples achieves \textasciitilde72\% accuracy and F1, which is lower than the values in the previous two tasks. The fine-tuned \texttt{RoBERTa-Twitter} \cite{pei-etal-2022-tweetfinsent} has similar performance. We also conduct an ablation study by removing emojis. Both \texttt{ChatGPT} and \texttt{GPT-4} show 2-3 points performance drop, indicating emojis in social media do convey meaningful sentiment signals. We do not have results of \texttt{Prior LLMs} as this dataset is not evaluated in the corresponding previous studies.

\begin{table}[htbp]
\centering
\resizebox{\columnwidth}{!}{%
\begin{tabular}{l|c|c}
\hline
\textbf{Model} & \textbf{Accuracy}  & \textbf{Weighted F1} \\
\hline
\texttt{ChatGPT{\scriptsize (0)}} & 68.48  & 68.60 \\
\texttt{ChatGPT{\scriptsize (5)}} & 69.93  & 70.05 \\
\texttt{GPT-4{\scriptsize (0)}} & 69.08  & 69.17 \\
\texttt{GPT-4{\scriptsize (5)}} & \underline{71.95}  & \textbf{72.12} \\ \hline
\texttt{ChatGPT({\scriptsize (0\_no\_emoji)})} & 64.40 & 64.43 \\ 

\texttt{ChatGPT({\scriptsize (5\_no\_emoji)})} & 67.37 & 67.61 \\

\texttt{GPT-4({\scriptsize (0\_no\_emoji)})} & 67.26 & 67.45 \\
\texttt{GPT-4({\scriptsize (5\_no\_emoji)})} & 70.58 & 70.44 \\ \hline
\texttt{RoBERTa-Twitter} & \textbf{72.30} & \underline{71.96} \\
\hline
\end{tabular}%
}
\caption{Results on the TweetFinSent dataset.}
\label{tab:TweetFinSent}
\end{table}

\subsection{Headline Classification}
While sentiment analysis has been regarded as one of the most basic tasks and is mainly pertaining to some dimensions of \textit{semantic orientation} \cite{osgood}, the semantics involved in financial text classification tasks can be more complicated.  
Classification, particularly multi-class text classification, is often applied to a wide range of financial text such as news, SEC 10-Ks, and market research reports to accelerate business operations. 

Same as in~\cite{wu2023bloomberggpt}, we use the news headlines classification dataset \cite{sinha2020impact} from the FLUE benchmark \cite{shah2022flue}. This classification task targets to classify commodity news headlines to one of the six categories like ``Price Up'' and ``Price Down''. We follow the setting in \texttt{BloombergGPT}, converting the multi-class classification to six individual binary classification problems (refer to Figure~\ref{Headlines_prompt} as an example). 

The model performance is listed in Table~\ref{tab:model_scores}. 
Again \texttt{GPT-4} outperforms \texttt{ChatGPT} and \texttt{Prior LLMs} as well as  \texttt{BloombergGPT}. 
The fine-tuned \texttt{BERT} can achieve 95\% on F1, 9\% higher than 5-shot \texttt{GPT-4}. This task is considered to be challenging due to its multi-class and the need of domain knowledge of the commodity market. 


\begin{table}[h]
\fontsize{10pt}{10pt}\selectfont
\centering
\begin{tabular}{l|c}
\hline
\textbf{Model}  & \textbf{Weighted F1} \\ \hline
\texttt{ChatGPT{\scriptsize (0)}} & 71.78 \\ 
\texttt{ChatGPT{\scriptsize (5)}}   & 74.84 \\ 
\texttt{GPT-4{\scriptsize (0)}}    & 84.17 \\
\texttt{GPT-4{\scriptsize (5)}}    & \underline{86.00} \\ \hline
\texttt{BloombergGPT{\scriptsize (5)}}  & 82.20 
\\ \hline
\texttt{GPT-NeoX{\scriptsize (5)}}  & 73.22 \\ 
\texttt{OPT66B{\scriptsize (5)}}     & 79.41 \\ 
\texttt{BLOOM176B{\scriptsize (5)}}   & 76.51 \\ \hline
\texttt{BERT}  & \textbf{95.36} \\ \hline
\end{tabular}
\caption{Results on the  headline classification task. }
\label{tab:model_scores}
\end{table}

\vspace{-1mm}
\subsection{Named Entity Recognition}
NER helps structure textual documents by extracting entities. It is a powerful technique to automate document processing and knowledge extraction from documents \cite{Yang.2021}. In our evaluation, we use the NER FIN3 datasets,  created by \citet{salinas-alvarado-etal-2015-domain} using financial agreements from SEC and containing four NE types: PER, LOC, ORG and MISC. Following the setting used in \texttt{BloombergGPT}, we remove all entities with the MISC label due to its ambiguity. 

In Table~\ref{Tab:NER_result}, we can see that both \texttt{GPT-4} and \texttt{ChatGPT} perform poorly under the zero-shot setup. Following \texttt{BloombergGPT}'s setting, the few-shot learning uses 20 shots on this dataset. We can see that \texttt{GPT-4} is less effective than \texttt{BloombergGPT}, and is comparable or worse than \texttt{Prior LLMs} on this task. Since NER is a classic structured prediction problem, \texttt{CRF} model is also compared. When \texttt{CRF} is trained with FIN5, which is similar to the test data (FIN3), it performs better than all the other models (see the last row of the table). 
Note that \texttt{CRF} is very sensitive to domain shifting---when it is trained on the out-of-domain CoNLL data, it performs poorly on the FIN3 data (refer to the second to the last row of Table~\ref{Tab:NER_result}), inferior to the zero-shot LLMs. In general, in this structured prediction task, LLMs' performance is not ideal and future improvement is imperative, particularly for the generalist models.

\begin{table}[htbp]\fontsize{10pt}{10pt}\selectfont
\centering
\begin{tabular}{l|c}
\hline
\textbf{Model}          & \textbf{Entity F1} \\ \hline
\texttt{ChatGPT{\scriptsize (0)}}        & 29.21      \\
\texttt{ChatGPT{\scriptsize (20)}}        & 51.52      \\
\texttt{GPT-4{\scriptsize (0)}}          & 36.08      \\
\texttt{GPT-4{\scriptsize (20)}}          & 56.71      \\ \hline
\texttt{BloombergGPT{\scriptsize (20)}}  & 60.82      \\ \hline
\texttt{GPT-NeoX{\scriptsize (20)}}       & \underline{60.98}      \\
\texttt{OPT66B{\scriptsize (20)}}         & 57.49      \\
\texttt{BLOOM176B{\scriptsize (20)}}      & 55.56      \\ \hline
\texttt{CRF{\scriptsize (CoNLL)}}          &  17.20      \\
\texttt{CRF{\scriptsize (FIN5)}}          &  \textbf{82.70}      \\
\hline
\end{tabular}
\caption{Results of few-shot performance on the NER dataset. \texttt{CRF{\scriptsize (CoNLL)}} refers to CRF model that is trained on general CoNLL data, \texttt{CRF{\scriptsize (FIN5)}}  refers to CRF model that is trained on FIN5 data. Again, we choose the same shot as \texttt{BloombergGPT} for fair comparison. More detailed experiments using 5 to 20 shots can be found in Appendix \ref{app:fewshot}.}
\label{Tab:NER_result}
\end{table}
\vspace{-1mm}

\subsection{Relation Extraction}
Ration extraction aims to detect linkage between extracted entities. It is a foundational component for knowledge graph construction, question answering and semantic search applications for the financial industry. In this study, we use a financial relation extraction dataset --- REFinD, which was created from 10-K/Q filings with 22 relation types \citet{kaur2023refind}. In order for  LLMs to predict the relationship between two entities, we provide the original sentence, entity words, and their entity types in the prompts and ask the models to predict a relation type. Same as in \texttt{Luke-base}~\cite{yamada-etal-2020-luke}, we use Macro F1. Table~\ref{Tab:REFinD} shows that the fine-tuned \texttt{Luke-base} outperforms both \texttt{ChatGPT} and \texttt{GPT-4} by a notable margin. On the other hand, \texttt{GPT-4} demonstrates considerably better performance compared to \texttt{ChatGPT}. The outcomes from this IE task illustrated the strength of fine-tuning on complex tasks that need a better understanding of the structure of sentences.


\begin{table}[htbp]
\fontsize{10pt}{10pt}\selectfont
\centering
\begin{tabular}{l|c}
\hline
\textbf{Model}   & \textbf{Macro F1} \\ \hline
\texttt{ChatGPT{\scriptsize (0)}}     & 20.97    \\
\texttt{ChatGPT{\scriptsize (10)}}     & 29.53    \\
\texttt{GPT-4{\scriptsize (0)}}       & 42.29   \\
\texttt{GPT-4{\scriptsize (10)}}       & \underline{46.87}   \\ \hline
\texttt{Luke-base{\scriptsize (fine-tune)}}   & \textbf{56.30} \\ \hline
\end{tabular}
\caption{Results on the REFinD dataset.}
\label{Tab:REFinD}
\end{table}

\vspace{-1mm}
\subsection{Question Answering}
The application of QA to finance presents a possible path to automate financial analysis, which at present is almost 100\% conducted by trained financial professionals. It is conventionally thought of as being challenging since it often requires a model to understand not only domain knowledge but also the embedded internal logic and numerical operation/reasoning. We adopt two QA datasets: FinQA \citep{chen2022finqa} and ConvFinQA \citep{chen2022convfinqa}. The former dataset focuses on a single question and answer pair. The latter decomposes the task into a multi-round structure: a chain of reasoning through conversation. Both of them concentrate on numerical reasoning in financial analysis, e.g. calculating profit growth ratio over years from a financial table. The experiment setting and prompt design details are in Appendix \ref{app:dataset_details} and \ref{app:fewshot}. Since the labels of the ConvFinQA test set are not publicly available, we utilize its dev dataset (421 samples) instead to evaluate the models, while for FinQA use the testing dataset (1,147 samples). 

\begin{table}[h]
\small
\vspace{0cm}
\centering
\begin{tabular}{l|c|c}
\hline
\textbf{Model} & \textbf{FinQA}  & \textbf{ConvFinQA} \\ \hline
\texttt{ChatGPT{\scriptsize (0)}}        & 48.56            & 59.86         \\
\texttt{ChatGPT{\scriptsize (3)}}        & 51.22            & /         \\
\texttt{ChatGPT{\scriptsize (CoT)}}      & 63.87             & /         \\
\texttt{GPT-4{\scriptsize (0)}}          &68.79 & \textbf{76.48}  \\
\texttt{GPT-4{\scriptsize (3)}}          & \underline{69.68}  & /  \\
\texttt{GPT-4{\scriptsize (CoT)}}        &\textbf{78.03}    &  / \\ \hline
\texttt{BloombergGPT{\scriptsize (0)}}   & /                & 43.41              \\ \hline
\texttt{GPT-NeoX{\scriptsize (0)}}       & /                & 30.06               \\ 
\texttt{OPT66B{\scriptsize (0)}}         & /                & 27.88               \\ 
\texttt{BLOOM176B{\scriptsize (0)}}      & /                & 36.31              \\ \hline
\texttt{FinQANet{\scriptsize (fine-tune)}} & 68.90   & \underline{61.24}         \\ \hline
\texttt{Human Expert}                     & 91.16             & 89.44            \\ 
\texttt{General Crowd}                    & 50.68             & 46.90           \\ \hline
\end{tabular}
\caption{Model performance (accuracy) on the question answering tasks. \texttt{FinQANet} here refers to the best-performing \texttt{FinQANet} version based on \texttt{RoBERTa-Large} \cite{chen2022finqa}. Few-shot and CoT learning cannot be executed on ConvFinQA due to the conservation nature of ConvFinQA.}
\label{tab:task1 result}
\end{table}

From the performance in Table~\ref{tab:task1 result}, we can see that \texttt{GPT-4} substantially outperforms all the other LLMs in both datasets.  For FinQA, \texttt{GPT-4} has highest zero-shot accuracy of 68.79\%, while  \texttt{ChatGPT} has 48.56\%. The performance gap between \texttt{GPT-4} and \texttt{ChatGPT} persists on ConvFinQA. \texttt{ChatGPT} has a big edge over \texttt{BloombergGPT} (59.86\% vs.~43.41\%) and also \texttt{Prior LLMs} on ConvFinQA. This result demonstrates that the continuous improvement of reasoning developed through \texttt{ChatGPT} to \texttt{GPT-4}, which is also observed in other studies.

We further explore the impact of few-shot learning and Chain-of-Thought (CoT) prompting on \texttt{GPT-4} and \texttt{ChatGPT}  on the FinQA task. The results provide a compelling narrative of performance increase using these prompting strategies. Both \texttt{ChatGPT} and \texttt{GPT-4} show a 1-3\% accuracy increase using 3 shots. This is consistent with our observations from other tasks. The CoT strategy brings a massive lift, 10\% and 15\% percentage points, to \texttt{ChatGPT} and \texttt{GPT-4} respectively. These results underscore the importance of detailed reasoning steps over shallow reasoning in boosting the performance of language models on complex financial QA tasks. The best \texttt{GPT-4} result indeed exceeds the fine-tuned FinQANet model with a quite significant margin. It is surprising to us since we previously observe that fine-tuned models have advantages on more complex tasks. We reckon that the scale of parameters and pre-training approaches make \texttt{ChatGPT} and \texttt{GPT-4} excel in reasoning than other models, particularly the numerical capability of \texttt{GPT-4}, which was demonstrated when the model was released by OpenAI.  But their performance (70+\% accuracy) still cannot match that of professionals (\textasciitilde90\% accuracy). Furthermore, numerical reasoning is just one of many reasoning tasks. More studies are needed for symbolic reasoning and other logic reasoning \cite{qin:23} if more datasets in the financial sector are further available. Also, we think the pretraining strategy such as RLHF has not been designed to improve sequence-labeling and structured-prediction skills needed in IE, but can inherently benefit QA.

\section{Discussions}


\paragraph{Comparison over LLMs.} We are able to benchmark the performance of \texttt{ChatGPT} and \texttt{GPT-4} with four other LLMs on five tasks with eight datasets. \texttt{ChatGPT} and \texttt{GPT-4} significantly outperforms others in almost all datasets except the NER task. It is interesting to observe that both models perform better on financial NLP tasks than \texttt{BloombergGPT}, which was specifically trained on financial corpora. This might be due to the larger model size of the two models. Finally, \texttt{GPT-4} constantly shows 10+\% boost over \texttt{ChatGPT} in straightforward tasks such as Headlines and FiQA SA. For challenging tasks like RE and QA, \texttt{GPT-4} can introduce 20-100\% performance growth. This indicates that \texttt{GPT-4} could be the first choice for financial NLP tasks before a more powerful LLM emerges. 


\paragraph{Prompt Engineering Strategies.} We adopted two commonly used prompting strategies: few-shot and chain-of-thoughts. We constantly observe 1\% to 4\% performance boost on \texttt{ChatGPT} and \texttt{GPT-4} from few-shot over zero-shot learning across various datasets. Chain-of-thoughts prompting is very effective in our test and demonstrates 20-30\% accuracy improvement over zero-shot and few-shot as well. According our findings, we argue that these two strategies should always be considered first when applying LLMs to financial NLP tasks. 

\paragraph{LLMs vs. Fine-tuning.} One attractive benefit of using LLMs in business domains is that they can be applied to a broad range of NLP tasks without conducting much overhead work. It is more economical compared to fine-tuning separate models for every task. Whereas, our experiments show fine-tuned models still demonstrate strong performance in most of the tasks except the QA task. 
Notably, for tasks like NER and RE, LLMs are less effective than fine-tuned models. In the QA tasks,
LLMs illustrated the advantage over fine-tuned model. But the reasoning complexity of the tested QA tasks is still deemed as basic in financial analysis. Although \texttt{ChatGPT} and \texttt{GPT-4} have proven to be able to perform multi-step reasoning, including numerical reasoning, to some extent, simple mistakes have still been made.


\paragraph{Using LLMs in Financial Services.} This study suggests that one can consider adopting the state-of-the-art generalist LLMs to address the relatively simple NLP tasks in financial applications. For more complicated tasks such as structured prediction, the pretraining plus fine-tuning paradigm is still a leading option. Although \texttt{ChatGPT} and \texttt{GPT-4} excel on QA compared to other models and are better than the general crowd, they are still far from satisfactory from the industry requirement standpoint. Significant research and improvement on LLMs are required before they can act as a trustworthy financial analyst agent. 

\section{Conclusion}
This study is among the first to explore the most recent advancement of \textit{generically} trained LLMs, including \texttt{ChatGPT} and \texttt{GPT-4}, on a wide range of financial text analytics tasks. These models have been shown to outperform models fine-tuned with domain-specific data on some tasks, but still fall short on others, particularly when deeper semantics and structural analysis are needed. While we provide comprehensive studies on eight datasets from five categories of tasks, we view our effort as an initial study, and further investigation of LLMs on financial applications is highly desirable, including the design of more tasks to gain further insights on the limitations of existing models, the integration of LLMs in the loop of human decision making, and the robustness of the models in high-stakes financial tasks.

\section*{Acknowledgement}

This research was funded in part by the Faculty Research Awards of J.P. Morgan AI Research. The authors are solely responsible for the contents of the paper and the opinions expressed in this publication do not reflect those of the funding agencies.
\\
\section*{Disclaimer}

This paper was prepared for informational purposes in part by the Artificial Intelligence Research group of JPMorgan Chase \& Co. and its affiliates ("JP Morgan''), and is not a product of the Research Department of JP Morgan. JP Morgan makes no representation and warranty whatsoever and disclaims all liability, for the completeness, accuracy or reliability of the information contained herein. This document is not intended as investment research or investment advice, or a recommendation, offer or solicitation for the purchase or sale of any security, financial instrument, financial product or service, or to be used in any way for evaluating the merits of participating in any transaction, and shall not constitute a solicitation under any jurisdiction or to any person, if such solicitation under such jurisdiction or to such person would be unlawful.

\bibliography{chatgpt}
\bibliographystyle{acl_natbib}

\clearpage
\appendix


\section{Details of the Related Work}
\label{app:related_work}
\paragraph{\texttt{ChatGPT} and Related Models.}
\texttt{ChatGPT}, \texttt{GPT-3.5} (\texttt{text-davinci-003}), and \texttt{GPT-4} are all part of a series of large language models created by OpenAI. \texttt{GPT-4}, as the latest and most advanced version, builds on the achievements of its forerunners. \texttt{ChatGPT} is an earlier version, tailored to offer users engaging and responsive conversational experiences. \texttt{GPT-3.5} acted as a transitional stage between \texttt{GPT-3} and \texttt{GPT-4}, improving upon the former and paving the way for the latter.

\texttt{ChatGPT} presents unprecedented quality when interacting with humans conversationally while retaining accumulated knowledge and generalization ability, achieved through large-scale conversational-style dataset pre-training and reward model fine-tuning. This allows \texttt{ChatGPT} to answer follow-up questions, admit mistakes, challenge incorrect premises, and reject inappropriate requests. Secondly, it is trained with a human-aligned objective function using Reinforcement Learning from Human Feedback (RLHF), which results in its output being more closely aligned with human preferences.

Recent studies have shown that \texttt{ChatGPT} outperforms multiple state-of-the-art zero-shot LLMs on various tasks and even surpasses fine-tuned models on some tasks. However, like many LLMs, \texttt{ChatGPT} still fails in many cases, such as generating overly long summaries or producing incorrect translations. A recent study \citep{bang:23} evaluated \texttt{ChatGPT}'s performance on multitasking, multilingual  and multimodal tasks, highlighting the importance of addressing these failure cases for improving the overall performance of the model.

\citet{qin:23} studied \texttt{ChatGPT}'s zero-shot capabilities on a diverse range of NLP tasks, providing a preliminary profile of the model. Their findings suggest that while \texttt{ChatGPT} shows certain generalization capabilities, it often underperforms compared to fine-tuned models on specific tasks. Compared to \texttt{GPT-3.5}, \texttt{ChatGPT} outperforms it on natural language inference, question answering, and dialogue tasks, while its summarization ability is inferior.  Both \texttt{ChatGPT} and \texttt{GPT-3.5} face challenges on sequence tagging tasks.

\paragraph{Domain-specific Models.}
Currently, there has been only a handful of financial-domain-specific LLMs available, which are often trained exclusively on domain-specific data. These LLMs have shown promising results in their respective domain tasks. For instance, \citet{Luo_2022} developed an LLM for the legal domain, which was trained exclusively on legal texts, and  \cite{taylor2022galactica} trained a healthcare LLM.

Most recently, \texttt{BloombergGPT} \citep{wu2023bloomberggpt}, a language model with 50 billion parameters, is trained using a mixed approach to cater to the financial industry's diverse tasks while maintaining competitive performance on general-purpose LLM benchmarks. A training corpus with over 700 billion tokens is created by leveraging Bloomberg's proprietary financial data archives and combining them with public datasets.  The model, designed based on the guidelines from \citep{hoffmann2022training} and \citep{scao2022language}, is validated on standard LLM benchmarks, open financial benchmarks, and Bloomberg-internal benchmarks. The mixed training approach results in a model that significantly outperforms existing models in financial tasks and performs on par or even better in some general NLP benchmarks.

It is worth mentioning that other researchers opt to adapt large general-purpose language models to tackle domain-specific tasks. For example, \citet{singhal2022large} applied \texttt{GPT-3} in the legal domain, and~\citet{lewkowycz2022solving} adapted \texttt{T5} to the financial domain. Despite being trained on a general purpose corpus, these models have also demonstrated excellent performance when applied to domain-specific tasks. Note that in addition to fine-tuning, research has also been conducted to use parameter efficient tuning for financial tasks such as intent detection on Banking77 dataset~\cite{li2022learning}.

\section{Dataset Details}
\label{app:dataset_details}

\paragraph{Financial PhraseBank.} This is a dataset introduced by \citet{malo2013good}, which is a sentiment classification dataset derived from financial news sentences. It is designed to assess the impact of news on investors, with positive, negative, or neutral sentiment labels being assigned to each news sentence from an investor's perspective. Containing 4,845 English sentences, the dataset is sourced from financial news articles found in the LexisNexis database. These sentences were annotated by individuals with expertise in finance and business, who were tasked with assigning labels based on their perception of the sentence's potential influence on the mentioned company's stock price.

\paragraph{FiQA Sentiment Analysis.} This second sentiment analysis task is part of the FiQA challenge \citep{10.1145/3184558.3192301} focusing on the prediction of sentiment specifically related to aspects within English financial news and microblog headlines. This was initially released as part of the 2018 competition that centered on financial question answering and opinion mining. The primary dataset was marked on a continuous scale, but we follow \texttt{BloombergGPT}'s setting and  transform it into a classification system with three categories: negative, neutral, and positive. We've created our own test split incorporating both microblogs and news. We use a 0-shot learning and our results are calculated through the weighted F1 score. We fine-tuned a RoBERTa-large model on this task for comparison with OpenAI and other LLMs.

\paragraph{TweetFinSent.} This third sentiment analysis task is introduced by \cite{pei-etal-2022-tweetfinsent}. The unique attribute of the TweetFinSent dataset is that it annotates tweets not merely on emotional sentiment, but also on the anticipated or realized gains or losses from a specific stock. Previous studies have revealed the TweetFinSent dataset as a challenging problem with significant room for improvement in the realm of stock sentiment analysis.

\paragraph{Headlines.} This binary classification task, created by \citet{sinha2020impact}, involves determining whether a news headline contains gold price related information. This dataset contains 11,412 English news headlines which span from 2000 to 2019. The headlines were collected from various sources, including Reuters, The Hindu, The Economic Times, Bloomberg, as well as aggregator sites. We note that the dataset we have access to consists of six tags:  ``price up'', ``price down'', ``price stable'', ``past price'', ``future price'', and ``asset comparison'', while the test reported in \texttt{BloombergGPT} used a version of nine categories. We contacted the original dataset authors, they claimed that they had performed some additional filtering and provided this six-label dataset. 

We also conducted an experiment where we prompted \texttt{ChatGPT} and \texttt{GPT-4} to generate answers simultaneously in response to six distinct questions. Our preliminary findings suggest that these models handle single-question prompts more effectively than those involving multi-tag binary classification. We noticed a significant drop in performance related to three tags: \textit{`past information'}, \textit{`future information'}, and \textit{`asset comparison'}. This suggests that the models struggle to provide separate and accurate responses to a series of questions presented at once. 

\paragraph{NER.} This named entity recognition task focuses on financial data collected for credit risk assessment from financial agreements filed with the U.S. Securities and Exchange Commission (SEC). The dataset, created by \citet{salinas-alvarado-etal-2015-domain}, consists of eight manually annotated documents with approximately 55,000 words. These documents are divided into two subsets: ``FIN5'' for training and ``FIN3'' for testing. The annotated entity types follow the standard CoNLL format \citep{tjong-kim-sang-de-meulder-2003-introduction} and include PERSON (PER), LOCATION (LOC), ORGANIZATION (ORG), and MISCELLANEOUS (MISC).

\paragraph{REFinD.} This relation extraction dataset is created by \citet{kaur2023refind}. REFinD is currently the most extensive of its kind, consisting of approximately 29K instances and 22 relations amongst 8 types of entity pairs. This specialized financial relation extraction dataset is constructed from raw text sourced from various 10-X reports (including but not limited to 10-K and 10-Q) of publicly traded companies. These reports were obtained from the website of the U.S. Securities and Exchange Commission (SEC).

\paragraph{ConvFinQA.} This is an extension of the FinQA dataset, named as ConvFinQA \citep{chen2022convfinqa}, which is designed to address numerical reasoning chains in a format of conversational question-answering tasks. ConvFinQA expands the original FinQA dataset to include 3,892 conversations with 14,115 questions derived from earnings reports of S\&P 500 companies. This task not only demands numerical reasoning and understanding of structured data and financial concepts, but also emphasizes the ability to relate follow-up questions to previous conversation context.

For the ConvFinQA dataset, we employ a turn-based approach, where we collect the answer generated by the models after each turn, append it to the previous question, and use them along with the next question as the prompt input for the next round. As shown in Figure~\ref{ConvFinQA_prompt}, we collect Answer 1 (A1) after Question 1 (Q1) and then prefix A1 together with Question 2 (Q2) to proceed to the next round, and so on, until we reach the end of the conversation chain. 

We notice some fundamental issues of \texttt{ChatGPT} from the tests on the ConvFinQA dataset. Firstly, it makes some basic mistakes, such as miscalculating ``\$753 million + \$785 million + \$1,134 million'' to be \$3,672 million instead of \$2,672 million. Even though all the intermediate results are correct, the final summation step produces an incorrect final answer. Note that such mistakes can be critical in the financial domain, particularly in high-stakes setups. We also found that \texttt{ChatGPT} struggles with understanding contextual information and coreference in conversations. For example, in ConvFinQA, questions often use the word ``that'' to refer to an entity mentioned in the previous question, but \texttt{ChatGPT} sometimes responds with a request for clarification, indicating its limitations in handling coreference. In contrast, \texttt{GPT-4} shows significant improvement and faces this issue much less.

\paragraph{FinQA.} \citet{chen2022finqa} propose an expert-annotated dataset consisting of 8,281 financial question-answer pairs, along with their corresponding numerical reasoning processes. Created by eleven finance professionals, FinQA is based on earnings reports from S\&P 500 companies~\cite{zheng2020global}. The questions necessitate extracting information from both tables and unstructured texts to provide accurate answers. The reasoning processes involved in answering these questions comprise common financial analysis operations, including mathematical operations, comparison, and table aggregation operations. FinQA is the first dataset of its kind designed to address complex question-answering tasks based on real-world financial documents.

When composing prompts, we use text and tables as context input, following the pattern `pre\_text' + `table' + `post\_text', where the `pre' and `post' texts provide the necessary context for the table, and the table itself contains the structured data that the model is expected to reason on and generate responses from. We also convert tables into a markdown format. For the FinQA dataset, we simply ask the question right after the context. Figure~\ref{FinQA_prompt} demonstrates the complete prompt format. 

We use the function call feature to assist CoT prompting. This Question\_Answering function required both models to generate two arguments: a) ``thinking process'' which contains each step of the reasoning process and evidence of how they locate information in the original documents and perform calculations, and b) ``answer'', which is the final numerical response.

We also conduct experiments with each of these models being subjected to different steps complexity, classified as 1-step programs, 2-step programs, and programs that involve more than 2 steps of calculation. For problems involving less than 2 steps, The models’ performance follows the same trend as overall results, where \texttt{GPT-4} maintains the lead, outperforming \texttt{FinQANet} and \texttt{ChatGPT}. However, the conclusion changes with the increase in problem complexity. When faced with problems requiring more than 2 steps, \texttt{ChatGPT} outperformed \texttt{FinQANet} by a significant margin, scoring accuracy of 32.14\% as opposed to \texttt{FinQANet}'s 22.78\%. It is intriguing to note that despite struggling with less complex tasks, \texttt{ChatGPT} managed to outpace \texttt{FinQANet} when problem complexity escalated. 

\begin{figure}[h]
    \centering
    \includegraphics[width=0.45\textwidth]{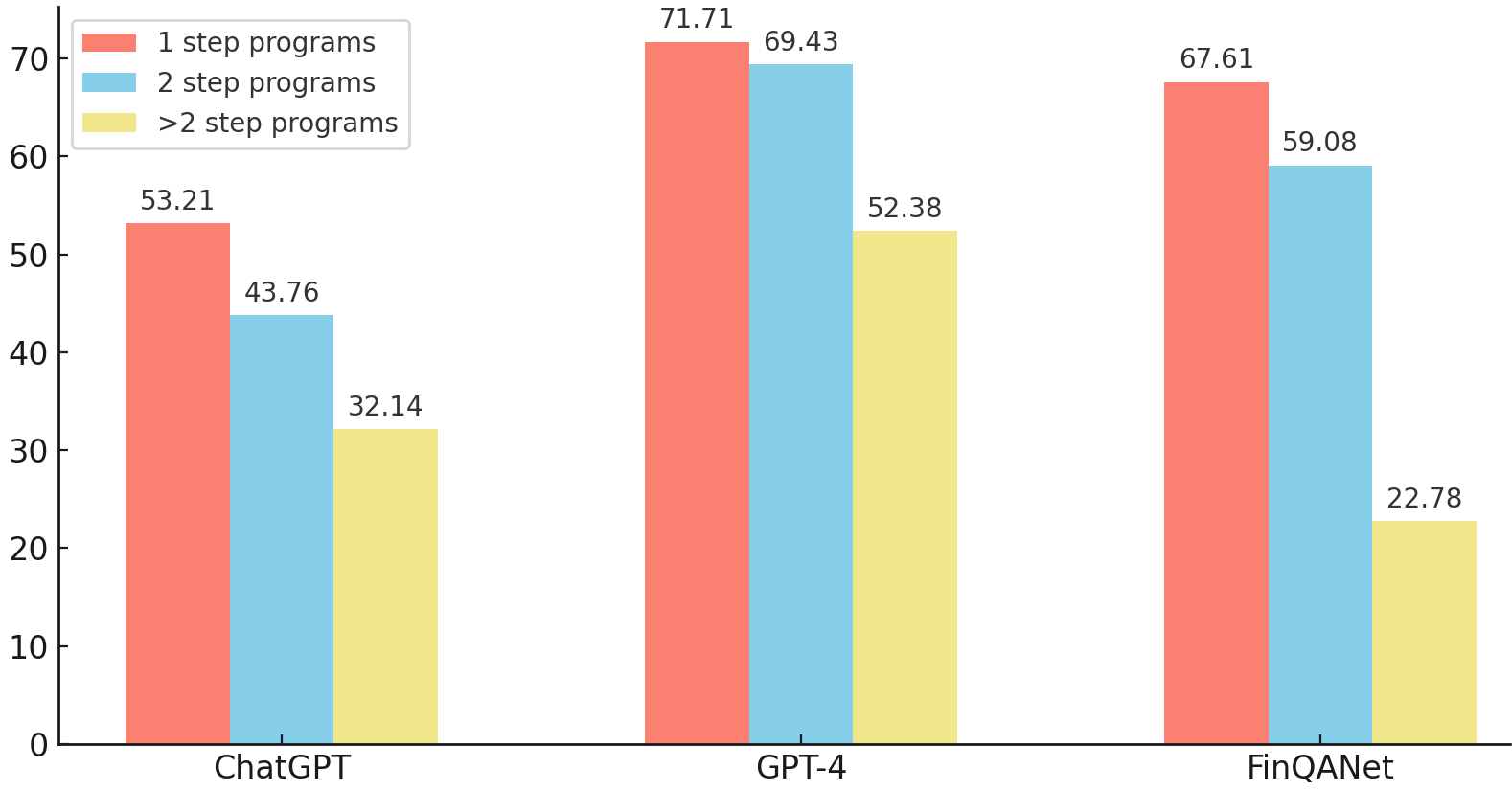}
    \caption{FinQA program steps analysis}
    \label{fig:NER}
\end{figure}

\section{Few Shots Experiments}
\label{app:fewshot}
We conducted few-shot experiments on 6 widely used datasets out of 8. We argue that ConvFinQA task itself is designed with a multi-step QA setup so we didn't conduct few-shot experiments on this dataset. For each shot number, we ran the experiment 10 times and generated box plots, which can be found in Figure~\ref{1} to ~\ref{5} below. The general trend shows that as we increase the number of shots, the performance of \texttt{ChatGPT} improves by approximately 1\% to 4\% across various datasets, in comparison to zero-shot. For simpler tasks, such as Sentiment Analysis (illustrated in Figure \ref{2}), \texttt{ChatGPT} only requires 6 shots to perform effectively. However, as we continue to increase the number of shots, the rate of improvement tapers off. For NER tasks, 5 shots do not impart sufficient domain information to \texttt{ChatGPT}, thus necessitating more than 15 shots to adequately guide the model. Additionally, we observed that performance can still fluctuate even with the same number of shots. The dispersion illustrated in the box plots indicates a certain level of volatility, suggesting that \texttt{ChatGPT} is quite sensitive to the shots used. This underlines the importance of careful selection and design of shots and prompts.

\begin{figure}[h]
    \centering
    \begin{minipage}{0.45\textwidth}
        \centering
        \includegraphics[width=1\textwidth]{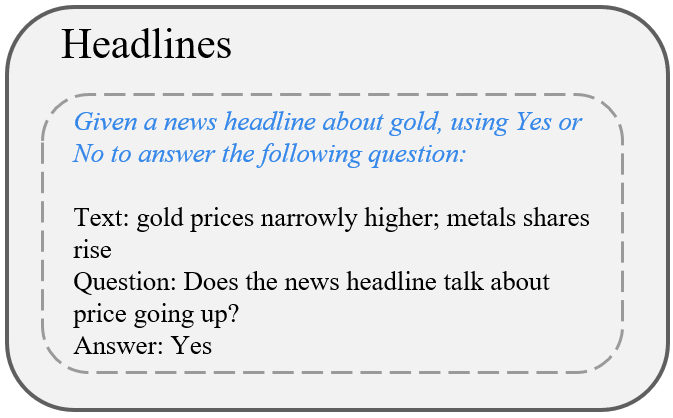}
        \caption{prompt for Headlines dataset}
        \label{Headlines_prompt}
    \end{minipage}\hfill
    \begin{minipage}{0.45\textwidth}
        \centering
        \includegraphics[width=1\linewidth]{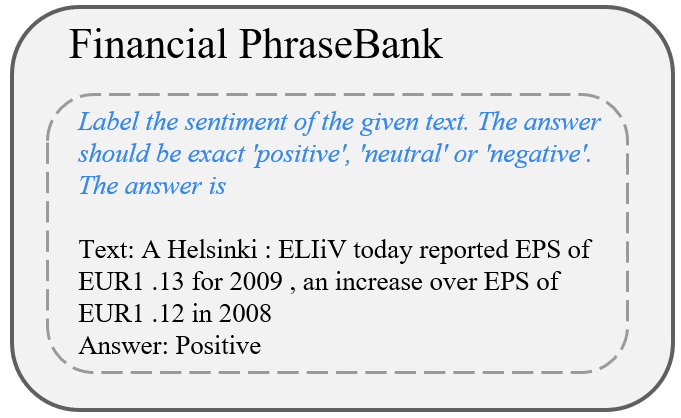}
        \caption{prompt for FPB dataset, same for other sentiment analysis tasks}
        \label{FPB_prompt}
    \end{minipage}
\end{figure}

\begin{figure}[h]
    \centering
    \begin{minipage}{0.45\textwidth}
        \centering
        \includegraphics[width=1\linewidth]{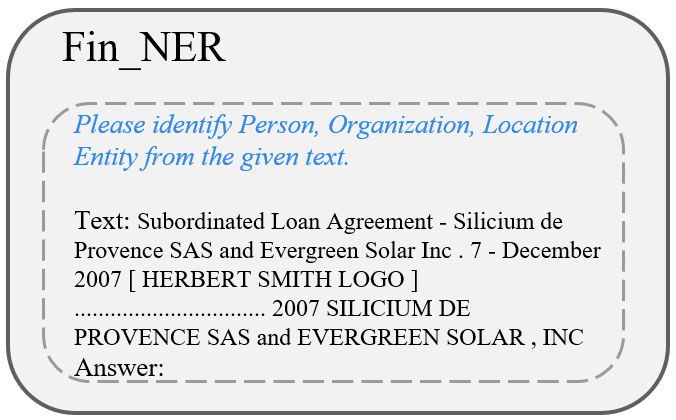}
        \caption{prompt for NER dataset}
        \label{NER_prompt}
    \end{minipage}\hfill
    \begin{minipage}{0.45\textwidth}
        \centering
        \includegraphics[width=1\linewidth]{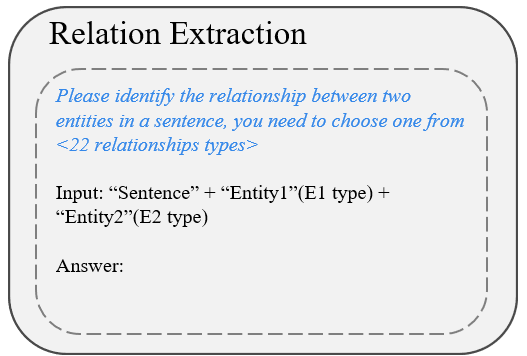}
         \caption{prompt for Relation Extraction dataset}
        \label{RE_prompt}
    \end{minipage}
\end{figure}

\begin{figure}[h]
    \centering
    \begin{minipage}{0.45\textwidth}
        \centering
        \includegraphics[width=1\linewidth]{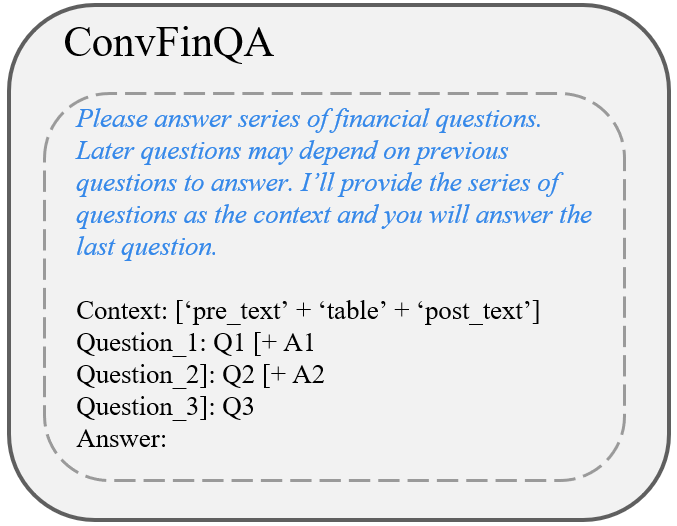}
        \caption{prompt for ConvFinQA dataset}
        \label{ConvFinQA_prompt}
    \end{minipage}\hfill
    \begin{minipage}{0.45\textwidth}
        \centering
        \includegraphics[width=1\linewidth]{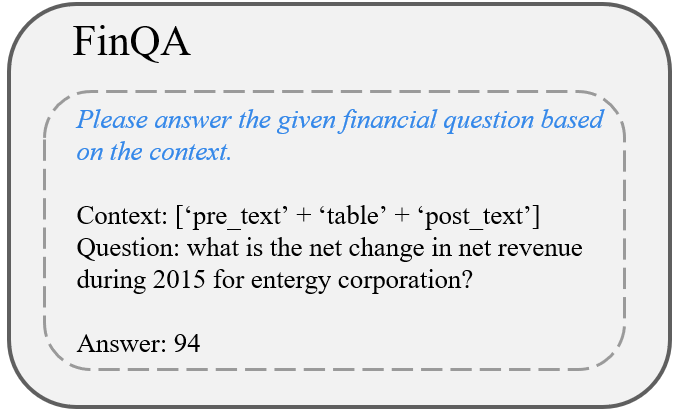}
        \caption{prompt for FinQA dataset}
        \label{FinQA_prompt}
    \end{minipage}
\end{figure}

We also listed the zero-shot prompt we used for each dataset, please find them in Figure~\ref{Headlines_prompt} to \ref{FinQA_prompt}. We use slightly different prompts for few-shot and CoT experiments since the shots and function call already provide guidance on how to structure the output.

\onecolumn 

\begin{table}[h]
\centering

\begin{tabular}{|l|l|}
\hline
Category & Tag Question \\\hline
price up & Does the news headline talk about price going up? \\
price stable & Does the news headline talk about price staying constant? \\
price down & Does the news headline talk about price going down? \\
past price & Does the news headline talk about price in the past? \\
future price & Does the news headline talk about price in the future? \\
asset comparison & Does the news headline compare gold with any other asset? \\\hline
\end{tabular}
\caption{Each tag and its corresponding converted question}
\label{tab:tagtoqa}
\end{table}

\begin{figure}[h]
    \centering
    \begin{minipage}{0.45\textwidth}
        \centering
        \includegraphics[width=1\linewidth]{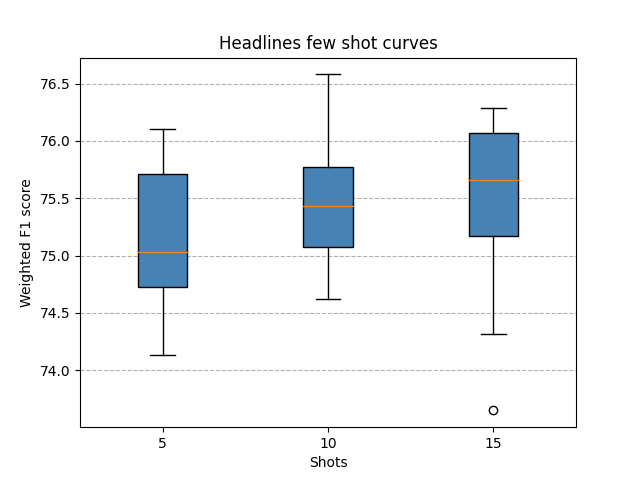}
        \caption{Headlines few shot results curve}
        \label{1}
    \end{minipage}\hfill
    \begin{minipage}{0.45\textwidth}
        \centering
        \includegraphics[width=1\linewidth]{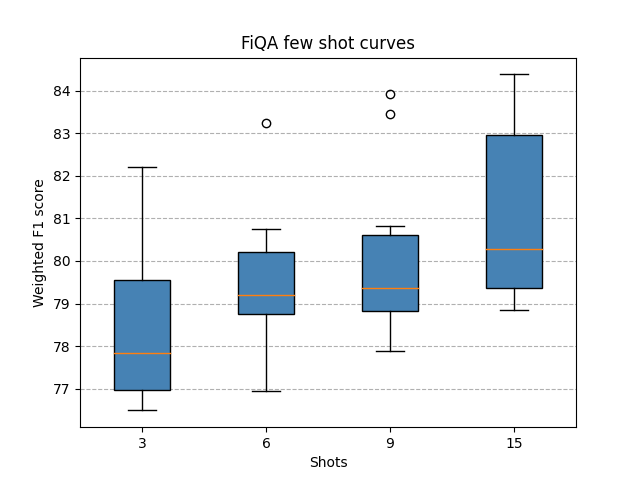}
        \caption{FiQA few shot results curve}
        \label{2}
    \end{minipage}
\end{figure}

\begin{figure}[h]
    \centering
    \begin{minipage}{0.45\textwidth}
        \centering
        \includegraphics[width=1\linewidth]{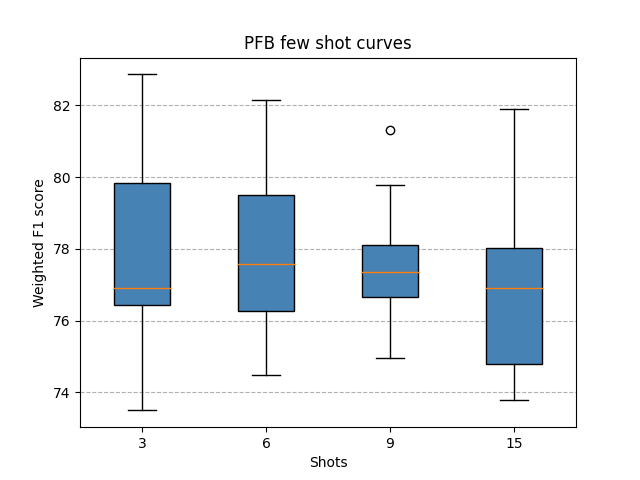}
        \caption{PFB few shot results curve}
        \label{3}
    \end{minipage}\hfill
    \begin{minipage}{0.45\textwidth}
        \centering
        \includegraphics[width=1\linewidth]{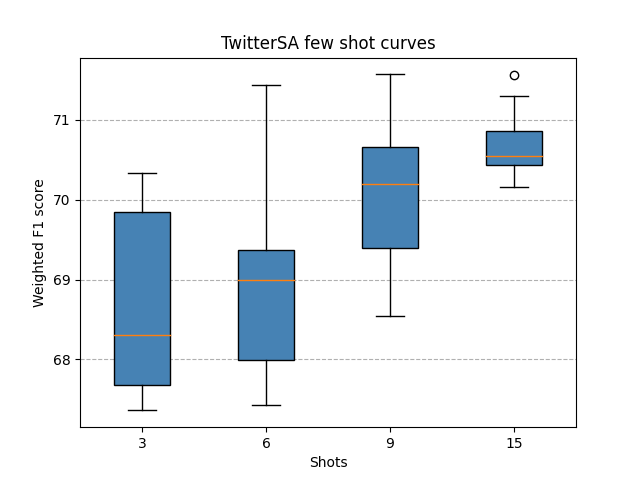}
        \caption{TweetFinSent few shot results curve}
        \label{4}
    \end{minipage}
\end{figure}

\begin{figure}[h]
    \centering
    \includegraphics[width=0.5\textwidth]{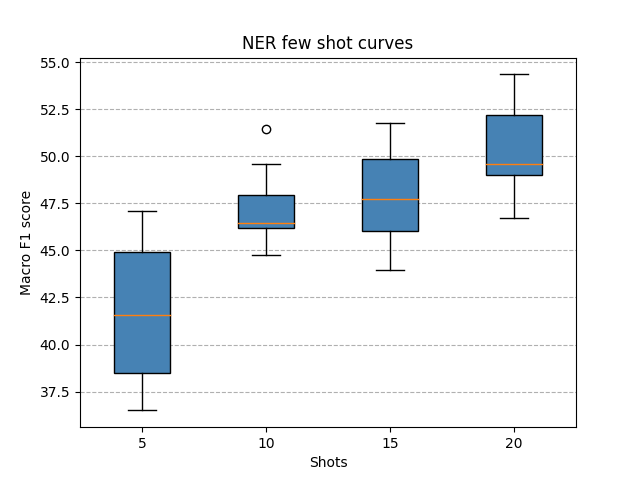}
    \caption{NER few shot results curve}
    \label{5}
\end{figure}

\end{document}